# PREVENTING OVERFITTING IN DEEP LEARNING USING DIFFERENTIAL PRIVACY

by

Alizishaan Anwar Hussein Khatri

1 SEPTEMBER 2017

A thesis submitted to the

Faculty of the Graduate School of

University at Buffalo, State University of New York

in partial fulfillment of the requirements for the

degree of

Master of Science

Department of Computer Science and Engineering

To the *'sulemaani keeda'* within - I couldn't have done this without you

To grad school or not, that is the question..

-Millenial Shakespeare

# TABLE OF CONTENTS

# LIST OF FIGURES

# LIST OF TABLES

# ACKNOWLEDGMENTS

This work would not have been possible without the support and almost super-human patience of my advisor, Prof. Marco Gaboardi. I would like to thank Dr Varun Chandola for making time to serve on the Thesis committee, my boss Dr Alex O'Connor for always being willing to help and the folks at Google for Tensorflow[1] (and free cloud credits). I would also like to thank the open-source community for TFLearn[8]

# ABSTRACT

The use of Deep Neural Network based systems in the real world is growing. They have achieved state-of-the-art performance on many image, speech and text datasets. They have been shown to be powerful systems that are capable of learning detailed relationships and abstractions from the data. This is a double-edged sword which makes such systems vulnerable to learning the noise in the training set, thereby negatively impacting performance. This is also known as the problem of *overfitting* or *poor generalization*. In a practical setting, analysts typically have limited data to build models that must generalize to unseen data. In this work, we explore the use of a differential-privacy based approach to improve generalization in Deep Neural Networks.

# CHAPTER 1
# INTRODUCTION

The primary goal of deep learning, and machine learning in general is to learn computational model(s) that learn from the training data and generalize well to unseen data. Deep learning allows computational models that are composed of multiple processing layers to learn representations of data with multiple levels of abstraction. These methods have dramatically improved the state-of-the-art in speech recognition, visual object recognition, object detection and many other domains such as drug discovery and genomics. Deep learning discovers intricate structure in large data sets by using the backpropagation algorithm to indicate how a machine should change its internal parameters that are used to compute the representation in each layer from the representation in the previous layer.[24]

With strutural innovations such as Convolutional Neural Networks, Recurrent Neural Networks and Long Short Term Memory Networks, the reach of neural networks has been extended even futher. The Stochastic Gradient Descent algorithm (SGD) [5], a computationally efficient version of the original Gradient Descent Algorithm [32][23] is normally used for training such networks. A standard training procedure typically involves multiple passes through the training data. The trainer has a limited say in the nature and number of representations / abstractions that the network learns. Many of these complex relationships could be a result of the noise specific to the data set. When evaluated on unseen data, these undesirable representations may cause the network to output faulty inferences. The phenomenon of fitting to the noise in the data, or fitting "too closely" to the dataset is called overfitting. In a neural network setting, a number of approaches have been developed to control overfitting. These include early stopping (stopping training in response to deteriorating performance on a holdout set) [31], soft weight sharing [28], weight penalties of various kinds such as L1 and L2 regularization [27] and Dropout [35]. However, these approaches primarily work either via structural modifications in the network and / or in other ways independent of the optimizing algorithm.

There has been a long history of adding random weight noise in the context of neural networks to improve generalization. However, there has been significantly less exploration of this in the context of Deep Neural networks and its myriad structural architectural variants. One of the notable efforts in this area, [26] presents a technique to improve optimization through the addition of annealed Gaussian noise to the gradients. The effects of random noise addition at node-level have been explored in depth in [17]. While both works have reported an increase in performance and an implicit decrease in overfitting, noise addition in both these cases is random in nature. This randomness implies that the noise added is independent to the training data set. In this work, we explore a differential privacy based noise addition mechanism to control overfitting which offers a fixed privacy guarantee with respect to the training data. This feature makes our noise addition approach data-set dependent to some extent, especially if privacy budgets are enforced.

Differential Privacy[9][15] is a strong privacy guarantee standard that aims to protect the privacy of individual records in a dataset and simultaneously lets users make useful inferences about the dataset as a whole. The goal of machine learning is to learn useful features from a dataset as a whole and utilize it to make inferences on unseen data. An important problem in machine learning is overfitting, which can be a result of learning too much from individual records in the data set. The complementary objectives makes the use of differential privacy in deep learning an interesting prospect. A notable attempt in applying differential privacy in the area of Machine Learning focusses on creating reusable holdout sets (The Thresholdout algorithm).[11] [12] A direct application of differential privacy in Deep Learning is the Differentially Private Stochastic Gradient Descent (DPSGD) algorithm, a variant of SGD has been proposed to train models while providing privacy guarantees on the training data.[2] Through a series of experiments, this thesis evaluates the performance of the DPSGD algorithm from the perspective of overfitting i.e. improving generalization.

# CHAPTER 2

# PRELIMINARIES

This chapter aims to provide a brief introduction to the fundamental concepts used in the thesis. Introducing the ideas of Differential Privacy, Deep Neural Networks, Generalization and related sub-concepts, it explores the interplay between them.

## 2.1 Differential Privacy

Differential privacy [15][14][10] is a privacy guarantee standard for algorithms on aggregate databases. It aims to ensure that an individual's privacy is protected but it is still possible to make some useful inferences about the data set as a whole.

### 2.1.1 Formal Definition

Let $\epsilon$ be a positive real number and $\mathcal{A}$ be a randomized algorithm that takes a dataset as input (representing the actions of the trusted party holding the data).

Let $\text{im}\mathcal{A}$ denote the image of $\mathcal{A}$. The algorithm $\mathcal{A}$ is $\epsilon$-differentially private if for all datasets $D_1$ and $D_2$ that differ on a single element (i.e., the data of one person), and all subsets $S$ of $\text{im}\mathcal{A}$,:

$$\Pr[\mathcal{A}(D_1) \in S] \leq e^\epsilon \times \Pr[\mathcal{A}(D_2) \in S], \tag{2.1}$$

where the probability is taken over the randomness used by the algorithm.

This original definition has since been extended to account for the possibility that plain $\epsilon$-differential privacy can be broken with probability $\delta$ [13]

*Definition.* A randomized mechanism $\mathcal{M} : \mathcal{D} \to \mathcal{R}$ with domain $\mathcal{D}$ and range $\mathcal{R}$ satisfies $(\epsilon, \delta)$-differential privacy if for any two adjacent inputs $d, d' \in \mathcal{D}$ and for any subset of outputs $\mathcal{S} \subseteq \mathcal{R}$ it holds that

$$\Pr[\mathcal{M}(d) \in \mathcal{S}] \leq e^{\epsilon} \Pr[\mathcal{M}(d') \in \mathcal{S}] + \delta \tag{2.2}$$

### 2.1.2 Noise Mechanism

A common paradigm for approximating a deterministic real-valued function $f : \mathcal{D} \to \mathcal{R}$ with a differentially private mechanism is via additive noise calibrated to $f$s sensitivity $S_f$, which is defined as the maximum of the absolute distance $|f(d) - f(d')|$ where $d$ and $d'$ are adjacent inputs. (The restriction to a real-valued function is intended to simplify this review, but is not essential.) DPSGD [2], the algorithm that this study is based on, uses a variant of the Gaussian noise mechanism is defined by:

$$\mathcal{M}(d) \triangleq f(d) + \mathcal{N}(0, S_f^2 . \sigma^2) \tag{2.3}$$

where $\mathcal{N}(0, S_f^2 . \sigma^2)$, is the normal (Gaussian) distribution with mean 0 and standard deviation $S_f \sigma$. A single application of the Gaussian mechanism to function $f$ of sensitivity $S_f$ satisfies $(\epsilon, \delta)$-differential privacy if $\delta \geq \frac{4}{5} \exp\left(-(\sigma \epsilon)^2 / 2\right)$ and $\epsilon < 1$ [15]

### 2.1.3 Qualitative Properties

According to one of the more exhaustive (and widely-cited) works on the subject [15], Differential Privacy offers the following benefits that we can leverage:

1. *Protection against arbitrary risks*, Differential Privacy not only offers protection on privacy of the individual record but also guards against re-identification.

2. *Automatic neutralization of linkage attacks*, including all those attempted with all past, present, and future datasets and other forms and sources of auxiliary information

3. *Quantification of privacy loss* Differential privacy is not a binary concept, and has a measure of privacy loss. This permits comparisons among different techniques answering critical questions like: for a fixed bound on privacy loss, which technique provides better accuracy? For a fixed accuracy, which technique provides better privacy?

4. *Composition.* The quantification of loss opens doors to the analysis and control of cumulative privacy loss over multiple computations. Understanding the behavior of differentially private mechanisms under composition enables the design and analysis of complex differentially private algorithms from simpler differentially private building blocks. This property is of interest to us as we focus on privacy across multiple computations.

5. *Group Privacy.* Differential privacy permits the analysis and control of privacy loss incurred by groups, such as families, employees of a particular company, members of a religion, etc. From the perspective of our application, it implies graceful degradation of privacy guarantees if a high degree of correlation is present in the input.

6. *Closure Under Post-Processing* Differential privacy is immune to post-processing: A data analyst, without additional knowledge about the private database, cannot compute a function of the output of a differentially private algorithm $\mathcal{M}$ and make it less differentially private. That is, a data analyst cannot increase privacy loss, either under the formal definition or even in any intuitive sense, simply by sitting in a corner and thinking about the output of the algorithm, no matter what auxiliary information is available.

### *2.1.4 Composition Theorems Of Differential Privacy*

#### Sequential composition

If we query an $\epsilon$-differential privacy mechanism $t$ times, and the randomization of the mechanism is independent for each query, then the result would be $\epsilon t$-differentially private. In the more general case, if there are $n$ independent mechanisms: $\mathcal{M}_1, \ldots, \mathcal{M}_n$ whose privacy guarantees are $\epsilon_1, \ldots, \epsilon_n$ differential privacy, respectively, then any function $g$ of them on a given dataset $D$: $g(\mathcal{M}_1(D), \ldots, \mathcal{M}_n(D))$ is ($\sum_{i=1}^{n} \epsilon_i$)-differentially private. [14]

#### Parallel composition

If the previous mechanisms are computed on disjoint subsets of the private database then the function $g$ would be $(\max_i \epsilon_i)$-differentially private.[14]

## 2.2 Deep Learning

Deep Learning or Deep Neural Networks are a class of machine learning tools. They are defined as Artificial Neural Networks with more than one hidden layer. These have achieved state-of-the-art-results on many standard data sets. The training process of DNNs involves making multiple passes through the data set. In each pass, gradients are computed on the training set (or a subset of it) using backpropagation and the hyperparameters are updated with the help of some optimizing algorithm. The objective is to minimize some Loss function $\mathcal{L}(\theta)$

Once the hyperparameters are tuned, unseen data is forward propagated through the network to make inferences. There are multiple variants of Deep Networks like Convolutional Neural Networks, Recurrent Neural Networks, LSTMs, etc. Privacy of training data and overfitting [35] are major issues being faced by these networks.

### 2.2.1 Deep Neural Networks

Deep Neural Networks 2.1 consists of multiple nodes *or neurons* organized in layers. The

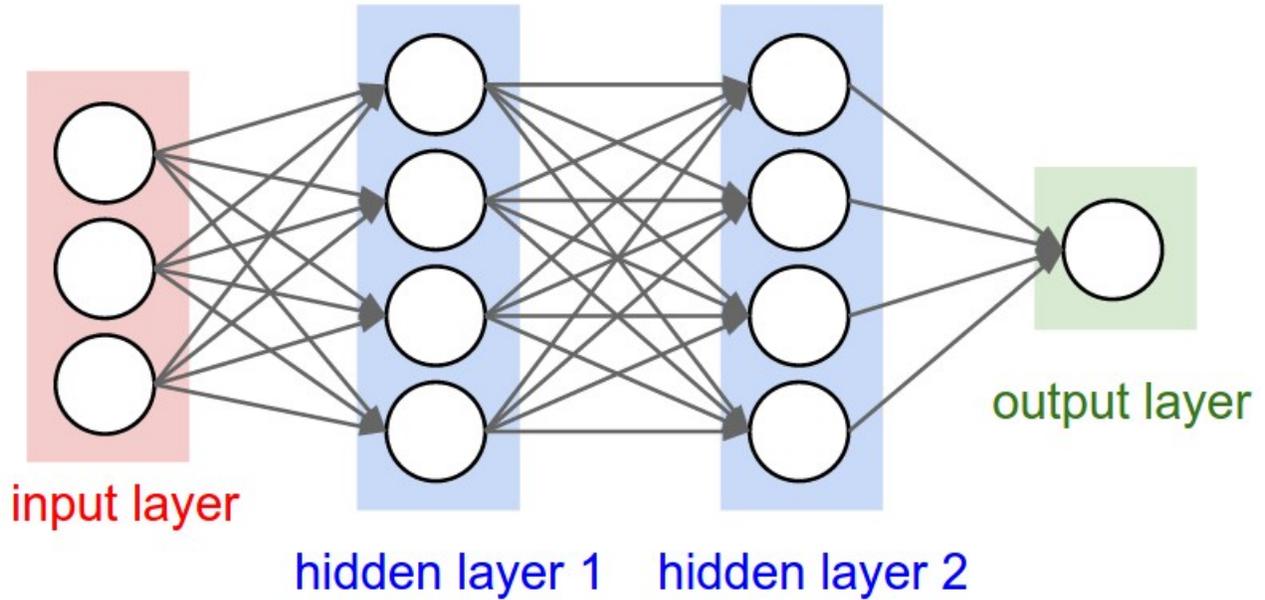

Figure 2.1: Deep Neural Network

outputs of neurons in layer $L_i$ are fed as input to the nodes in layer $L_{i+1}$ and so on. Optionally, each neuron might also be connected to an additional bias neuron. These links have weights which are learned during the training process. Given a neuron $i$ in layer $k$ with activation function $f$, its output $y_{i,k}$ is given by:

$$y_i^k = f(x) = f\left(\sum_{j=0}^{N} w_{i,j}^k y_j^{k-1}\right) \qquad (2.4)$$

where, $w_{i,j}^k$ is the weight of the link connecting neuron $i$ in layer $k-1$ to neuron $j$ in layer $k$

$N$ is the total number of neurons in layer $k-1$

Basically each neuron implements the function:

$$y = wx + b \tag{2.5}$$

where, $w$ is the weight-matrix and $b$ is the bias node

### 2.2.2 Rectified Linear (ReLU) activation function

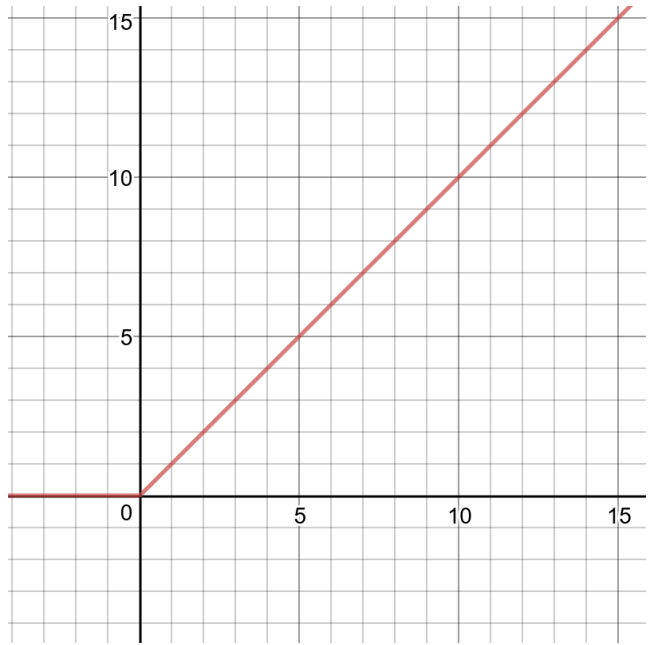

Figure 2.2: The Rectified Linear 'ReLU' Activation function

The Rectified Linear (ReLU) activation function gets its name from Half-wave Rectifiers in electronics, which it is analogous in behaviour to. It was first introduced to dynamic networks with strong biological and mathematical motivations.[18]( Ref Fig 2.2) In the context of artificial neural networks, the ReLU activation function is given by Equation 2.6

$$f(x) = x^+ = \max(0, x) \tag{2.6}$$

The intuition behind using it as the activation for our application is two fold: a) It has produced state-of-the-art results in multiple Deep Neural Network settings.[19][37][25] b) The outputs vary linearly with the input; this ensures that the added noise doesn't sway the values too much.

### 2.2.3 Softmax Activation function

The Softmax activation function, or normalized exponential function, is a generalization of the logistic function that "squashes" a mvar—K-dimensional vector **z** of arbitrary real values to a mvar—K-dimensional vector $\sigma(\mathbf{z})$ of real values in the range [0, 1] that add up to 1. The function is given by Equation 2.7

$$\sigma(\mathbf{z})_j = \frac{e^{z_j}}{\sum_{k=1}^{K} e^{z_k}} \tag{2.7}$$

for $j = 1,...,K$

A key property of this function is that it is differentiable which makes it learnable via *Back-propogation* and thereby, are good to be used in a Deep Neural Network settings.These softmax activations functions are often used to represent a categorical probability distribution over $j$ distinct classes.

### 2.2.4 Categorical Crossentropy loss function

This is the loss function that the training process seeks to minimize. It involves the computation of cross-entropy beweem the true and the approximating distribution.

Let $y_{true}^{(k)}$ and $y_{pred}^{(k)}$ be the true and predicted distributions respectively for a given output class $k$. Over $N$ samples, the loss function $L$ is given by equation 2.8.

$$L = -\sum_{i=1}^{N} y_{true}^{(1)} \log y_{pred}^{(1)} \tag{2.8}$$

For $K$-output classes, the cross entropy loss function is given by Eq. 2.9

$$L = -\sum_{i=1}^{N}\sum_{k=1}^{K} y_{true}^{(k)} \log y_{pred}^{(k)} \tag{2.9}$$

### 2.2.5 Backpropagation

Backpropagation is a method to calculate the gradient of the loss function with respect to the weights in an artificial neural network. It is commonly used as a part of algorithms that optimize the performance of the network by adjusting the weights, for example in the gradient descent algorithm. It is also called backward propagation of errors.

The optimization algorithm repeats a two phase cycle, propagation and weight update. When an input vector is presented to the network, it is propagated forward through the network, layer by layer, until it reaches the output layer. The output of the network is then compared to the desired output, using a loss function, and an error value is calculated for each of the neurons in the output layer. The error values are then propagated backwards, starting from the output, until each neuron has an associated error value which roughly represents its contribution to the original output.

Backpropagation uses these error values to calculate the gradient of the loss function. In the second phase, this gradient is fed to the optimization method, which in turn uses it to update the weights, in an attempt to minimize the loss function.

The importance of this process is that, as the network is trained, the neurons in the intermediate layers organize themselves in such a way that the different neurons learn to recognize different characteristics of the total input space. After training, when an arbitrary input pattern is present which contains noise or is incomplete, neurons in the hidden layer of the network will respond with an active output if the new input contains a pattern that resembles a feature that the individual neurons have learned to recognize during their training. It is a generalization of the delta rule to multi-layered feedforward networks, made possible by using the chain rule to iteratively compute gradients for each layer. Backpropagation requires that the activation function used by the artificial neurons (or nodes) is differentiable.

### 2.2.6    Differentially Private Stochastic Gradient Descent

Standard stochastic gradient descent helps the classifier learn very detailed representations of the training data efficiently and very quickly. [5] This is a double-edged sword as the learnt abstractions or representations of the data can be desirable or undesirable. For example, fitting to the noise in the data set can make the system produce undesirable results in test values with a similar underlying distribution but different noise distribution. This is why it becomes important that the network either learns only the desirable abstractions in the data set or learns an optimium balance of both. In other words, it can be said that the network must not learn the training data set 'too closely'. Adding noise in a controlled manner to the data set intuitively seems like a viable option to circumvent this issue. The next section explores this approach in detail.

## 2.3    Differentially Private Stochastic Gradient Descent

The idea to use a differentially private mechanism with Stochastic gradient Descent was first proposed in [34] It has also been followed in [3][2]

We use the approach given in [2] because of their simplicity and state-of-the-art Privacy

**Algorithm 1:** Differentially private SGD

**Data:** Examples $\{x_1, ..., x_N\}$
loss function $\mathcal{L}(\theta) = \frac{1}{N}\sum_i \mathcal{L}(\theta, x_i)$
Parameters : learning rate $\eta_t$, noise scale $\sigma$, group size $L$, gradient norm bound $C$
**Initialize** $\theta_0$ randomly
**for** $t \, \epsilon \, [T]$ **do**
    Take a random sample $L_t$ with sampling probability $L/N$
    **Compute Gradient**
    For each $i \, \epsilon \, L_t$, compute $g_t(x_i) \leftarrow \nabla_{\theta_t} \mathcal{L}(\theta, x_i)$
    **Clip gradient**
    $\bar{g}_t(x_i) \leftarrow g_t(x_i)/max\left(1, \frac{\|g_t(x_i)\|_2}{C}\right)$
    **Add noise**
    $\bar{g}_t \leftarrow \frac{1}{L}(\sum_i \bar{g}_t(x_i) + \mathcal{N}(0, \sigma^2 C^2 \mathbf{I}))$
    **Descent**
    $\theta_{t+1} \leftarrow \theta_t - \eta_t \bar{g}_t$
**Output** $\theta_\mathcal{T}$ and compute the overall privacy cost $(\epsilon, \delta)$ using a privacy accounting method.

accounting. They argue that adding noise during the training process is better than adding it to the final learnt parameters as the latter can have an adverse impact on the model in the worst case.

Algorithm 1 describes their approach in detail.

Algorithm 1 proposes three fundamental additions to the Standard Stochastic Gradient Descent algorithm.

Algorithm 1 outlines our basic method for training a model with parameters $\theta$ by minimizing the empirical loss function $\mathcal{L}(\theta)$. At each step of the SGD, the gradient $\nabla_\theta \mathcal{L}(\theta, x_i)$ is computed for a random subset of examples, the $\ell_2$ norm of each gradient is clipped, average is computed, noise is added in order to protect privacy, and a step is taken in the opposite direction of this average noisy gradient. At the end, in addition to out-putting the model, the algorithm also computes the net privacy loss of the mechanism based on the information maintained by the privacy accountant. The key components of the algorithm 1 are described in detail.

### 2.3.1 Gradient Clipping

Proving the differential privacy guarantee of Algorithm 1 requires bounding the influence of each individual example on $\bar{g}_t$. Each gradient is clipped in $\ell_2$ norm; i.e., the gradient vector $g$ is replaced by $g/max(1, \frac{\|g\|_2}{C})$ , for a clipping threshold C. This clipping ensures that if $\|g\|_2 \leq C$, then $g$ is preserved, whereas if $\|g\|_2 > C$, it gets scaled down to be of norm C. We remark that gradient clipping of this form is a popular ingredient of SGD for deep networks for non-privacy reasons[29], though in that setting it usually suffices to clip after averaging.

### 2.3.2 Lots

Like the ordinary SGD algorithm, DPSGD estimates the gradient of $\mathcal{L}$ by computing the gradient of the loss on a group of examples and taking the average. This average provides an unbiased estimator, the variance of which decreases quickly with the size of the group. We call such a group a lot, to distinguish it from the computational grouping that is commonly called a batch. In order to limit memory consumption, we may set the batch size much smaller than the lot size $L$, which is a parameter of the algorithm. We perform the computation in batches, then group several batches into a lot for adding noise. In practice, for efficiency, the construction of batches and lots is done by randomly permuting the examples and then partitioning them into groups of the appropriate sizes. For ease of analysis, however, we assume tat each lot is formed by independently picking each eample with probability $q = L/N$, where $N$ is the size of the input dataset. As is common in the literature, we normalize the running time of a training algorithm by expressing it as the number of epochs, where each epoch is the (expected) number of batches required to process $N$ examples. In our notation, an epoch consists of $N/L$ lots.

### 2.3.3 Privacy Accounting

The authors of [2] have presented an innovative privacy accounting mechanism called Moments accountant. For the Gaussian noise that we use, if $\sigma$ in DPSGD is chosen to be $\sqrt{2\log\frac{1.25}{\delta}}/\epsilon$ then by standard arguments [15] each step is $(\epsilon, \delta)$-differentially private with respect to the lot. Since the lot itself is a random sample from the database, the privacy amplification theorem [4] [21] implies that each step is $(O(q\epsilon), q\delta)$-differentially private with respect to the full database where $q = L/N$ is the sampling ratio per lot and $\epsilon \leq 1$. The result in the literature that yields the best overall bound is the strong composition theorem [16]. However, the strong composition theorem can be loose, and does not take into account the particular noise distribution under consideration. The Moments Accountant used shows that DPSGD is $(O(q\epsilon\sqrt{T}), \delta)$- differentially private for appropriately chosen settings of the noise scale and the clipping threshold. Compared to what one would obtain by the strong composition theorem, our bound is tighter in two ways: it saves a $\sqrt{log(\frac{1}{\delta})}$ factor in the $\epsilon$ part and a $Tq$ factor in the $\delta$ part. Since we expect $\delta$ to be small and $T \gg \frac{1}{q}$ (i.e., each example is examined multiple times), the saving provided by this bound is quite significant.

## 2.4 Generalization

Generalization refers to the ability of a learning algorithm to be able to learn the properties of the data that it has been trained on, not the data itself. As per the traditional definition [7], a learning algorithm can be said to *generalize* if, trained on a training set drawn i.i.d. from an underlying distribution, returns a hypothesis whose empirical error (on the training data) is close to its true error (on the underlying distribution). We deliberately overfit the hypothesis on the training set and then evaluate the learning algorithm's performance on freshly generated data from the same distribution to quantify generalization.

It might also be possible for an adversary to map the outputs of the learnt hypothesis to another vector space (possibly even the input vector space) with fair accuracy.[7]. This is further discussed in Section 5.1

***Definition*((Traditional) Generalization)**. Let $\mathcal{X}$ be an arbitrary domain. A mechanism $\mathcal{M} : \mathcal{X}_L^n \to (\mathcal{X} \to \{0, 1\})$ is $(\alpha, \beta)$-generalizing if for all distributions $\mathcal{D}_L$ over $\mathcal{X}_l$, given a sample $S_L \sim_{i.i.d.} \mathcal{D}_L^n$,

$$\Pr[\mathcal{M}(S_L) \text{ outputs } h : \mathcal{X} \to \{0,1\} \text{ such that } |err(h) - err(S_L, h)| \leq \alpha] \geq 1 - \beta$$

where the probability is over the choice of the sample $S_L$ and the randomness of $\mathcal{M}$.

## 2.5 Overfitting

Deep Neural Networks, like a majority of other Machine Learning systems known to man, are susceptible to overfitting. In the real world, data sets consist of both wanted and unwanted features or noise. When a learning algorithm is run on a training set, it learns both desirable and undesirable features. As the training process continues, the learning algorithm updates the hypothesis to mirror the training set more and more closely. In this process it learns desirable features as well as noise. When the hypothesis has learnt more than the optimum level of noise, its performance on unseen data begins to fall. This is when overfitting is said to have occurred. In a Neural Network setting, networks have been observed to overfit in one region of the input domain while underfitting in another.[22][6]. The complexity of Deep Neural Networks makes it difficult to understand overfitting in them when compared to most other machine learning methods.

A number of methods have been developed specifically to control overfitting in Deep Neural Networks. These include stopping the training in response to deteriorating performance on a validation set, introducing weight penalties of various kinds such as $l1$ and $l2$ Regularization, Soft weight sharing[32] and Dropout[35]. These techniques, notably Dropout, produce good performance in the primary objective of preventing overfitting in Deep Neural Network settings. These approaches primarily work either via structural modifications in the network. An alternate approach to prevents overfitting comes in the form of perturbing gradients in the optimization stage. A notable work in this area [26], focuses on adding annealed Gaussian noise to the gradients. Their noise addition strategy, however, is data set agnostic. Since, we use a differential-privacy based approach, we have the added provision to regulate noise addition tightly with respect to the training data set.

## 2.6 Summary

This chapter introduced the concepts of Differentially Privacy, Deep Learning, Generalization and Stochastic Gradient Descent (Private and Non-private). Highlighting the limitations of state-of-the-art methods, it asserts the need for a different approach that addresses them. It also presents a road map for the proposed solution.

# CHAPTER 3
# OUR APPROACH

This chapter describes the data set and network architectures used for the experiments in this paper. Here, we analyze some of the objectives of our thesis and develop techniques to achieve them. It also aims to provide the reader some insight behind the some of the design choices.

## 3.1 Modus Operandi

The traditional notion of generalization from classical Learning Theory explains overfitting as a function of the difference between the performance of the hypothesis on seen and unseen data. In order to demonstrate the efficiency of our approach, we hand craft a neural network and data set pair that achieves almost perfect accuracy on the training data but achieves a high error on unseen data. We then apply our technique and demonstrate a significant reduction in the differential error on training and test sets.

## 3.2 Dataset

### 3.2.1 Rationale

As described in Section 2.5, overfitting can be understood as a phenomenon where the learning mechanism produces a hypothesis that learns the noise in the data set very closely thereby negatively impacting its performance on the underlying data distrbution. Learning algorithms that are more robust to noise in the training data perform better on unseen data. A data set with a large noise component thus becomes a natual choice to evaluate the noise robustness of our learning algorithm.

### 3.2.2  Description

Our data sets consists of $n$ records containing 200 attributes each. All points are drawn from a Bernoulli distribution with a given success probability $p$. Equation 3.1 represents the distribution of each attribute in the data set. This distribution represents the noise in the data. Records are alternately assigned labels $l = 1$ and $l = -1$ *(later one-hot encoded to [0, 1] and [1, 0])*. Bias is inserted into the data set by replacing the last 100 attributes with points drawn from a Bernoulli Distribution with probability $p\sim$.*(See Equation 3.2)* Our data set is a highly noisy one where 50% of the attributes are noisy while the remaining 50% containig the class bias.

$$\Pr(\mathcal{X} = 1) = p = 1 - \Pr(\mathcal{X} = 0) \tag{3.1}$$

$$p\sim = p \,*\, (0.5 \pm b) \quad \text{where, b is the bias offset} \tag{3.2}$$

## 3.3  Network architecture

### 3.3.1  Keeping things simple

Deep Neural Networks can turn into very complex systems. It is very difficult to understand what abstractions are learnt internally by a neural network. Our primary objective here is to build a network which can learn both the underlying class and noise distributions of the training data. Neural Networks are very robust to overparametrization. Networks with higher model complexity than needed have been shown to generate similar results to their simpler counterparts.[22] [6] [20]. This suggests that neural networks that overfit on one model will overfit similarly on models with higher model complexity, unless there is some significant structural modification. The inactive nodes or neurons in high complexity models simply become inactive during the training process, thereby producing results similar to the optimal low complexity model. Training networks with high complexity requires significantly larger amounts of computational resources in comparision. These are the two major reasons why we were inclined to chose to use a simple neural network structure to begin with.

### 3.3.2 Details

We use a fully connected neural network with bias nodes for binary classification. Our net consists of an input layer, two hidden layers and an output layer. The first and second layers contain 128 and 16 nodes respectively. Both use the 'Rectified Linear' activation functions[19][18][37][25]. The output layer consists of two nodes using 'Softmax' activation.

For each record in the data set, let the input bits be $a_1, a_2, a_3, ...., a_{200}$. These are fed to the network via the input layer.

The output $y_i^{(1)}$ for each neuron $i$ where $i = 1, 2.., 128$ in hidden layer 1 is given by 3.3

$$y_i^{(1)} = \sum_{k=1}^{200} w_{i,k}^{(1)} a_k + b_i^{(1)} \quad (3.3)$$

where, $w_{i,k}^{(1)}$ is the weight of the link between input $a_k$ and neuron $i$; $b_i^{(1)}$ is the *bias* of the $i^{th}$ node in the $1^{st}$ hidden layer.

The outputs of this layer are fed as inputs to the second hidden layer.

The output $y_i^{(2)}$ for each neuron $i$ where $i = 1, 2.., 16$ in hidden layer 2 is given by Equation 3.4

$$y_i^{(2)} = \sum_{k=1}^{128} w_{i,k}^{(2)} y_k^{(1)} + b_i^{(2)} \tag{3.4}$$

All outputs from hidden layer two become inputs to the Softmax layer, which also is the output layer of the neural network. The number of neurons in this layer is equal to the number of classes that the classifier is to be trained to predict. Here, since we are predicting two classes, we use two neurons in the softmax layer.

The output of neuron $i$ for $i \in 1, 2$ is given by Equation 3.5

$$\sigma(\mathbf{z})_i = \frac{e^{z_i}}{\sum_{k=1}^{2} e^{z_k}} \tag{3.5}$$

In the training phase, we use the categorical crossentropy loss function.
Equation 3.6 represents the loss function used for the current two binary classfication problem. It is obtained by substituting $K = 2$ in Equation 2.9

$$L = -\sum_{i=1}^{N} \sum_{k=1}^{2} y_{true}^{(k)} \log y_{pred}^{(k)} \tag{3.6}$$

For binary classification (positive / negative class formulation), this can be rewritten as Equation 3.7

$$L = -\sum_{i=1}^{N} \left( y_{true} \log (y_{pred}) + (1 - y_{true}) \log (1 - y_{pred}) \right) \tag{3.7}$$

## 3.4 Summary

This chapter describes our motivations and approach to improving generalization in Deep Neural Networks. It also describes the construction of a data set and a network model and briefly discusses the motivations for doing so.

# CHAPTER 4

# QUANTIFYING GENERALIZATION

This chapter focuses on utilizing the traditional notion of generalization to analyse overfitting. It describes an experiment to quantify the generalizing behaviour of Stochastic Gradient Descent and Differentially Private Stochastic Gradient Descent algorithms. It also presents a comparative analysis of the results.

## 4.1 The Traditional Notion of Generalization

According to Classical Learning Theory[36], the process of learning begins with the assumption that there exists a distribution from the input space $X$ to the output space $Y$. A learning algorithm aims to learn an approximate hypothesis of this underlying distribution looking at a labelled set of $(input, output)$ pairs. It does so by minimizing the error over these training examples, also called *empirical error*. Traditionally, the efficiency of this hypothesis is evaluated by measuring the difference of accuracy (or error) over the training set and unseen test data. This generalizing behaviour can be quantified in a standardized dataset-agnostic fashion in terms of the parameters $\alpha$ and $\beta$. Re-writing the definition from Section 2.4:

**Definition** *Traditional Generalization* Let $\mathcal{X}$ be an arbitrary domain. A mechanism $\mathcal{M} : \mathcal{X}_L^n \to (\mathcal{X} \to \{0,1\})$ is $(\alpha, \beta)$-*generalizing*, if for all distributions $\mathcal{D}_L$ over $\mathcal{X}_L$, given a sample $\mathcal{S}_{L \sim i.i.d.} \mathcal{D}_L^n$,

$$\Pr[\mathcal{M}(\mathcal{S}_L) \text{ outputs } h : \mathcal{X} \to \{0,1\} \text{ such that } |err(h) - err(\mathcal{S}_L, h)| \leq \alpha] \geq 1 - \beta, \quad (4.1)$$

*where the probability is over the choice of the sample $\mathcal{S}_L$ and the randomness of $\mathcal{M}$.*

$\alpha$ can be understood to be the maximum differential error and and $\beta$ is the maximum anti-probability of $\alpha$, or the maximum probability of getting errors greater than $\alpha$.

## 4.2 Purposeful Overfitting

As described in Section 4.1, we utilize the traditional notion of generalization as a measure of overfitting. An overfit hypothesis is expected to perform very well on training data but poorly on fresh data or in other words, the gap between the performance of such a hypothesis on training and test sets will be very large. Section 4.3 describes an experiment that evaluates the generalization behavior of hypothesis generated using Stochastic Gradient Descent and Differentially Private Stochastic Gradient Descent.

## 4.3 Experimental Evaluation

### *4.3.1 Data Set*

We use a data set as described in Section 3.2.2 with $10^6 = 1000000$ records. The data set is divided into 10 bins with an equal number of records with 0 and 1 labels per bin. The classifier is independently trained on each one of the bins and tested on all 10 bins.

### *4.3.2 Procedure*

For each training session we calculate the error on the training set($train_{error}$), error on the test set ($test_{error}$) and the absolute value of the difference between them, $|train_{error} - test_{error}|$. We then iterate $\alpha$ in the range $0-1$ and calculate the minimum corresponding $\beta$ in each case using Equation 4.1.

The above steps are then repeated for data sets with different *noise scale($\sigma$)* values.

## 4.4  Results and Discussions

Figures 4.1, 4.2, 4.3 and 4.4 were generated on the data set described in section 4.3.1 using Equation 4.1. Each plot illustrates the relationship between $\alpha$ and $\beta$ for a fixed noise scale $\sigma$, for both SGD and DPSGD using the values in Tables 4.1, 4.2, 4.3 and 4.4 respectively. Each row in the table contains the results of training on a separate fold. For each row in the table, the first three columns correspond to the values generated using SGD and the next three columns contain values generated using DPSGD. For both algorithms, these three columns represent the error on the training set, the error on the entire distribution (train + test) and the absolute difference between these two values. $\alpha$ is plotted on the X-axis while $\beta$ is plotted on the Y-axis. We use the following values of hyper-parameters: *learning rate($\eta_t$)=0.1, gradient norm bound($C$) = 4.0, epochs=150, bias offset ($\tilde{p}$) = 0.05, $p = 0.5$, Lot Size($L$) = 960, Epochs = 150*

In all cases, we observe that the hypothesis generated by Stochastic Gradient Descent (SGD) produces an error of 0% & approx. 36% on the training data and test data respectively. This difference in error rates on training and fresh data is evidence of the classifier encoding the training data within itself. We have thus successfully recreated the conditions that take place during over-fitting due to data reuse.

The hypothesis generated by Differentially Private Stochastic Gradient Descent (DP-SGD) for different values of $\sigma$ exhibit similar performance on the training and test sets. The difference in errors has been reduced by about 66%!

As we increase the value of $\sigma$, the absolute errors on training and test sets begin to increase. This is explained by taking a look at the noise addition mechanism of DPSGD. The Normal Distribution from which the noise added begins to grow horizontally as $\sigma$ grows. This implies that the likelihood of drawing a point with a high magnitude grows. Adding a large magnitude of noise makes the value of the gradients diverge further away. In some training runs, gradient-explosion [30] too has been observed. This causes the learning algorithm to output zero hypothesis and thus in a data set with equal class representation between datasets, the performance on train is equivalent to 0.5 and difference in performance is close to 0%

## 4.5   Interpreting the results in terms of $\alpha$ and $\beta$

From figures 4.1, 4.2, 4.3 and 4.4, we observe that DPSGD lowers the maximum differential error, $\alpha_{max}$ by at least half. In Table 4.2, we observe very similar Test errors for DPSGD and SGD with the differential error $\alpha$ being approximately halved. More precisely, the maximum differential error ($\alpha_{max}$)has been reduced by about 60% while producing similar results on the test set. As $\sigma$ is increased further, i.e. noise is added from a larger distribution (lower / tighter privacy budget $\epsilon$), the error of the classifier on both the training and fresh data goes up significantly (Tables 4.1, 4.2, 4.3 and 4.4). While this is axiomatic due to the nature of the noise addition, an interesting observation here is that the differential error $\alpha$ is reduced in comparison to standard *Stochastic Gradient Descent*.

For the hypothesis generated using DPSGD, the area under the $\alpha$-$\beta$ curve has been reduced. While the co-domain of $\beta$ is similar in both cases, the maximum value of differential error, $\alpha_{max}^{dpsgd}$ has been reduced to a quantity less than half of the non-private value, $\alpha_{max}^{sgd}$. The probability of getting smaller differential errors is high while that of getting larger er-

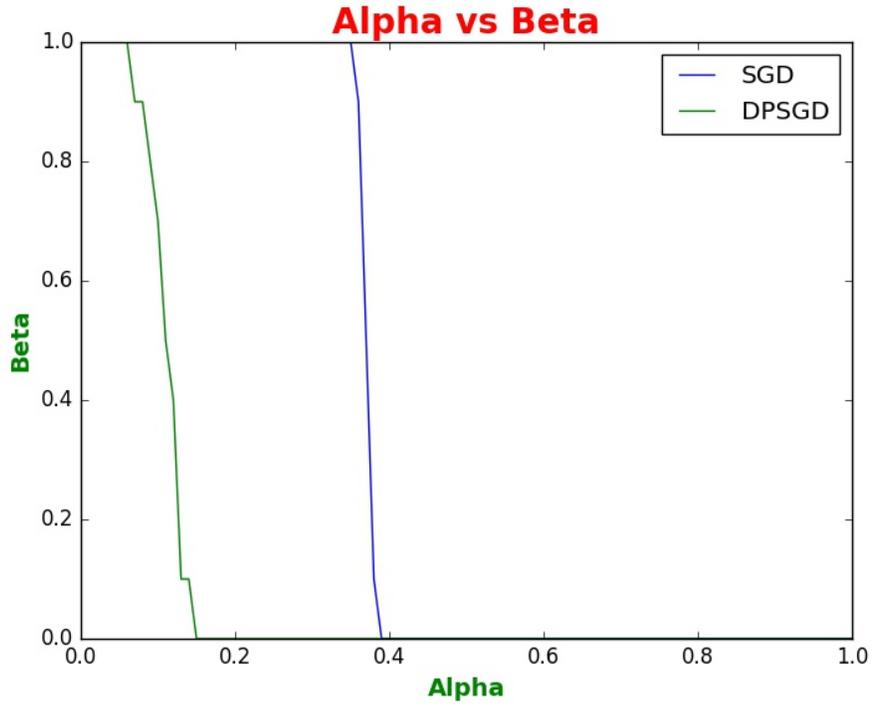

| SGD | | | DPSGD | | |
|---|---|---|---|---|---|
| train | test | $|train-test|$ | train | test | $|train-test|$ |
| 0.0 | 0.3631 | 0.3631 | 0.299 | 0.4059 | 0.1069 |
| 0.0 | 0.3873 | 0.3873 | 0.385 | 0.4457 | 0.0607 |
| 0.0 | 0.3751 | 0.3751 | 0.313 | 0.4158 | 0.1028 |
| 0.0 | 0.3673 | 0.3673 | 0.342 | 0.433 | 0.091 |
| 0.0 | 0.3698 | 0.3698 | 0.27 | 0.3962 | 0.1262 |
| 0.0 | 0.3754 | 0.3754 | 0.272 | 0.3973 | 0.1253 |
| 0.0 | 0.3737 | 0.3737 | 0.295 | 0.4201 | 0.1251 |
| 0.0 | 0.3584 | 0.3584 | 0.3 | 0.4114 | 0.1114 |
| 0.0 | 0.3769 | 0.3769 | 0.35 | 0.4311 | 0.0811 |
| 0.0 | 0.3694 | 0.3694 | 0.238 | 0.3852 | 0.1472 |

(a) Generalization values for $\sigma = 2.0$

Figure 4.1: $\alpha, \beta$-generalization plot for $\sigma = 2.0$

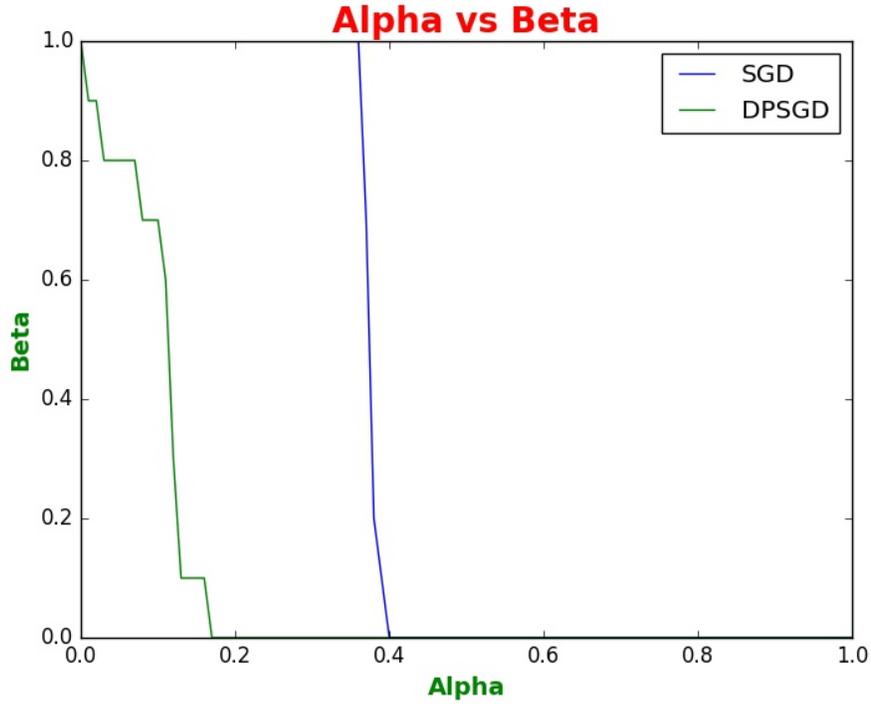

| SGD | | | DPSGD | | |
|---|---|---|---|---|---|
| train | test | $|train-test|$ | train | test | $|train-test|$ |
| 0.0 | 0.3798 | 0.3798 | 0.293 | 0.4114 | 0.1184 |
| 0.0 | 0.376 | 0.376 | 0.291 | 0.398 | 0.107 |
| 0.0 | 0.3689 | 0.3689 | 0.236 | 0.4053 | 0.1693 |
| 0.0 | 0.3754 | 0.3754 | 0.461 | 0.4894 | 0.0284 |
| 0.0 | 0.3804 | 0.3804 | 0.297 | 0.413 | 0.116 |
| 0.0 | 0.3742 | 0.3742 | 0.291 | 0.4057 | 0.1147 |
| 0.0 | 0.3799 | 0.3799 | 0.284 | 0.4101 | 0.1261 |
| 0.0 | 0.3673 | 0.3673 | 0.273 | 0.3999 | 0.1269 |
| 0.0 | 0.393 | 0.393 | 0.502 | 0.5001 | 0.0019 |
| 0.0 | 0.3699 | 0.3699 | 0.316 | 0.3934 | 0.0774 |

(a) Generalization values for $\sigma = 4.0$

Figure 4.2: $\alpha, \beta$-generalization plot for $\sigma = 4.0$

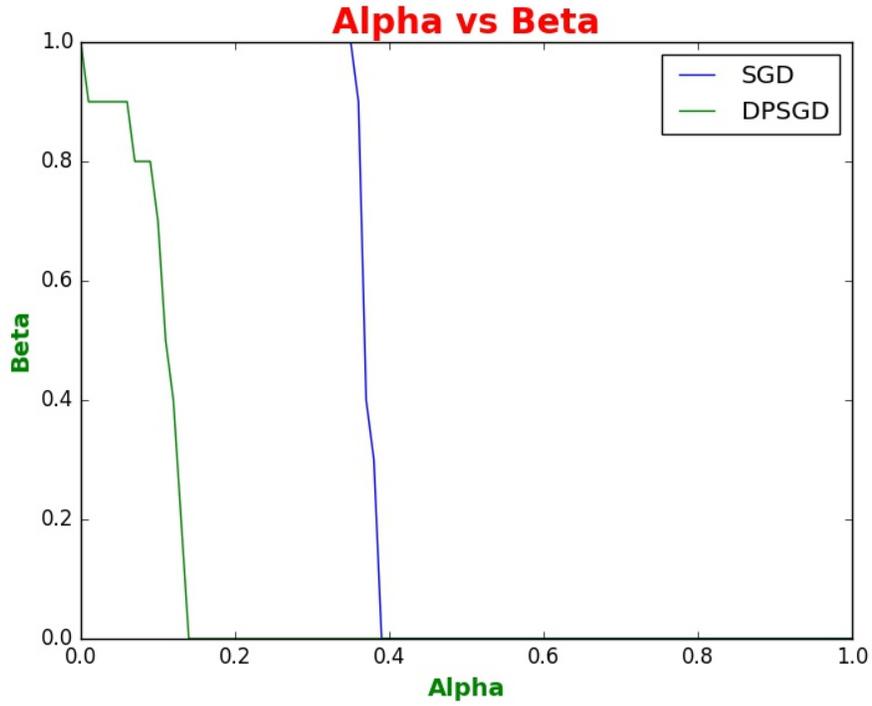

| SGD | | | DPSGD | | |
|---|---|---|---|---|---|
| train | test | $|train-test|$ | train | test | $|train-test|$ |
| 0.0 | 0.3777 | 0.3777 | 0.3 | 0.4296 | 0.1296 |
| 0.0 | 0.3667 | 0.3667 | 0.282 | 0.4159 | 0.1339 |
| 0.0 | 0.369 | 0.369 | 0.27 | 0.3973 | 0.1273 |
| 0.0 | 0.3887 | 0.3887 | 0.311 | 0.4173 | 0.1063 |
| 0.0 | 0.3804 | 0.3804 | 0.293 | 0.4047 | 0.1117 |
| 0.0 | 0.3868 | 0.3868 | 0.313 | 0.4161 | 0.1031 |
| 0.0 | 0.3669 | 0.3669 | 0.506 | 0.4961 | 0.0099 |
| 0.0 | 0.3617 | 0.3617 | 0.311 | 0.4039 | 0.0929 |
| 0.0 | 0.3586 | 0.3586 | 0.384 | 0.4487 | 0.0647 |
| 0.0 | 0.3661 | 0.3661 | 0.288 | 0.427 | 0.139 |

(a) Generalization values for $\sigma = 8.0$

Figure 4.3: $\alpha, \beta$-generalization plot for $\sigma = 8.0$

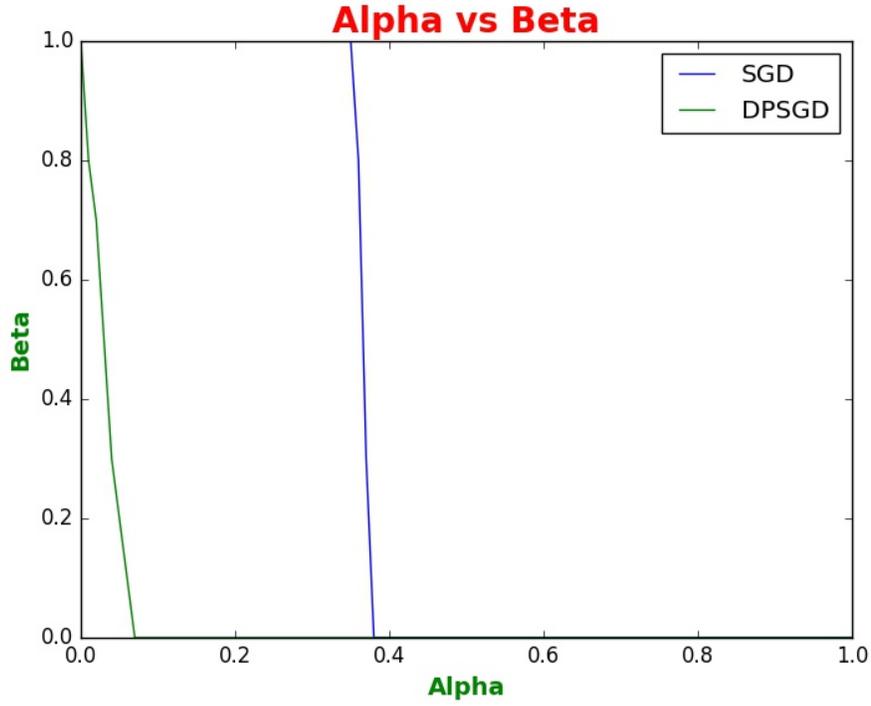

| SGD | | | DPSGD | | |
|---|---|---|---|---|---|
| train | test | $|train-test|$ | train | test | $|train-test|$ |
| 0.0 | 0.3661 | 0.3661 | 0.435 | 0.4638 | 0.0288 |
| 0.0 | 0.3624 | 0.3624 | 0.448 | 0.4952 | 0.0472 |
| 0.0 | 0.3634 | 0.3634 | 0.479 | 0.49 | 0.011 |
| 0.0 | 0.3726 | 0.3726 | 0.43 | 0.4851 | 0.0551 |
| 0.0 | 0.355 | 0.355 | 0.476 | 0.5023 | 0.0263 |
| 0.0 | 0.3606 | 0.3606 | 0.429 | 0.4627 | 0.0337 |
| 0.0 | 0.3679 | 0.3679 | 0.486 | 0.4918 | 0.0058 |
| 0.0 | 0.3723 | 0.3723 | 0.433 | 0.465 | 0.032 |
| 0.0 | 0.3797 | 0.3797 | 0.392 | 0.4528 | 0.0608 |
| 0.0 | 0.3581 | 0.3581 | 0.494 | 0.4956 | 0.0016 |

(a) Generalization values for $\sigma = 40.0$

Figure 4.4: $\alpha, \beta$-generalization plot for $\sigma = 40.0$

rors has been reduced. The probability of getting errors larger than $\alpha_{max}^{dpsgd}$ has been brough down to zero. Since, $\alpha_{max}^{dpsgd} \approx 0.4 \left(\alpha_{max}^{sgd}\right)$, the top 60% of the higher value errors have been eliminated. We can see that the probability of getting smaller differential errors has been increased at the cost of higher differential errors. We have also eliminated differential errors greater than $\alpha_{max}^{dpsgd}$.

In Table 4.2, we observe that the training errors for SGD and DPSGD differ by approx. 25 points, the latter quantity being the greater one. The test errors are similar. We can thus conclude that differentially private SGD produces similar test set performance while fitting significantly less closely too the training set. This is one of our key results.

## 4.6 Summary

This chapter presented an experiment to compare the generalizing behaviour of SGD and DPSGD learning algorithms. Giving a brief insight into the generalizing behaviour of the hypothesis, it reported a reduction of about 60% in max differential errors (with similar performance on unseen test data) and about 25 % point increase in training error while achieving similar performance on the test data.

# CHAPTER 5

# OVERFITTING

This chapter focuses on overfitting due to data reuse. It presents a detailed comparison of the performance of the hypothesis generated by Stochastic Gradient Descent (SGD) and Differentially Private Stochastic Gradient Descent (DPSGD) algorithms as a function of training epochs.

## 5.1 Overfitting due to data reuse

It is axiomatic that excessive learning iterations over the training data lead to overfitting the training data. While trying to estimate the optimal hypothesis for a given training sample, it is tricky to determine what construes 'too much'. There always is a risk that the classifier would learn both useful information and noise from the training data. Furthermore, the learning algorithm might simply encode the entire training data set in the hypothesis. This leaves the trainer (usually a human) to decide how much is too much. This could get tricky in practise as it involves balancing the trade-off between learning as much as possible from the data set and maintaining good generalization. We describe an experiment to perform a comparative analysis of SGD and DPSGD with respect to overfitting and convergence as the number of training steps (or epochs) are increased.

## 5.2 Convergence

As described in Section 2.5, overfitting is said to have happened when the learning algorithm produces a hypothesis which fits the noise distribution more closely than the underlying data distribution. We interpret this to mean that the classifier learns a hypothesis that converges more towards the training data as a whole than it does towards the underlying data distribution. In terms of convergence, overfitting is said to have occurred when the learning mechanism produces a hypothesis which converges more to the hybrid distribution of noise and the underlying data as opposed to converging to the underlying data distribution itself. This however is an ideal case as it might be impractical for a Deep Neural network to isolate the underlying distribution using empirical data, completely disregard the noise and fit only to underlying distribution. In a practical setting, a perfect fit is the point at which the partial convergences to the training data and underlying distribution are closest to each other.

## 5.3 Experimental Evaluation

### 5.3.1 Data Set

We start off with using the data set described in Section 3.2.2. The model is trained with 1000 train values and 1000 test values. We present results obtained by varying $\sigma$(noise scale) and *clipping threshold or gradient norm bound*.

### 5.3.2 Procedure

The data set is split equally into two sets (training and test). The classifier is trained on the training set and tested on the test set. This repeated independently for both SGD and DPSGD algorithms. The number of training epochs is varied and classifier performance is recorded at fixed intervals. We then plot a graph of the classifier's performance on each one of these sets as a function of the number of training epochs.

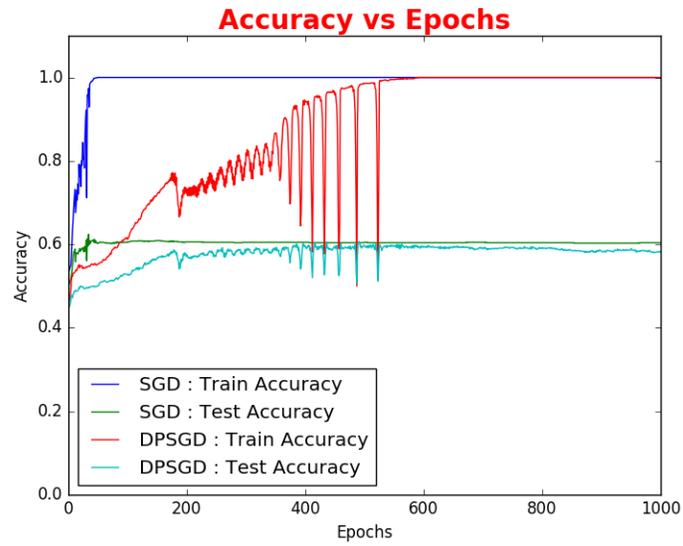

Figure 5.1: Classification accuracy as a function of epochs ($\sigma = 1.0$)

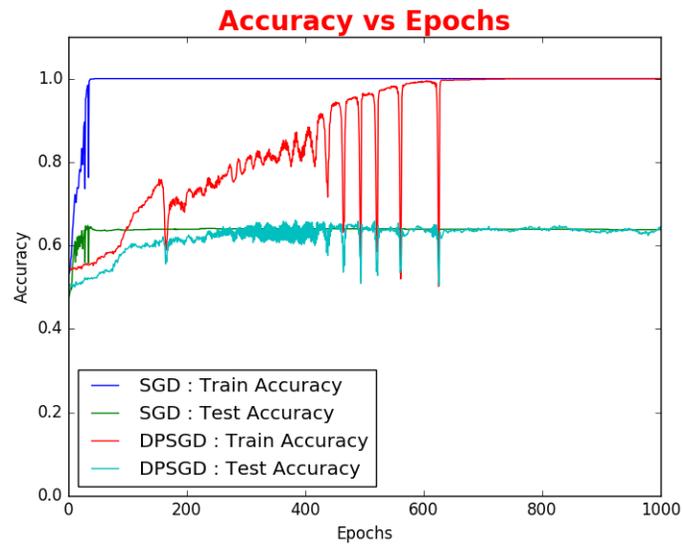

Figure 5.2: Classification accuracy as a function of epochs ($\sigma = 4.0$)

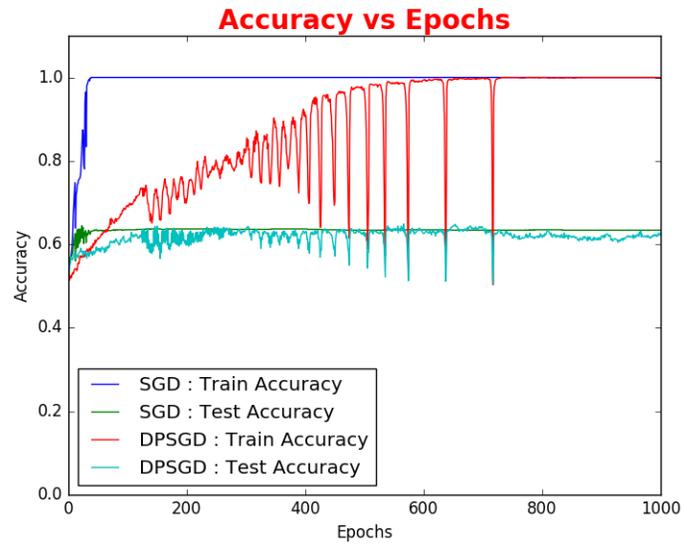

Figure 5.3: Classification accuracy as a function of epochs ($\sigma = 8.5$)

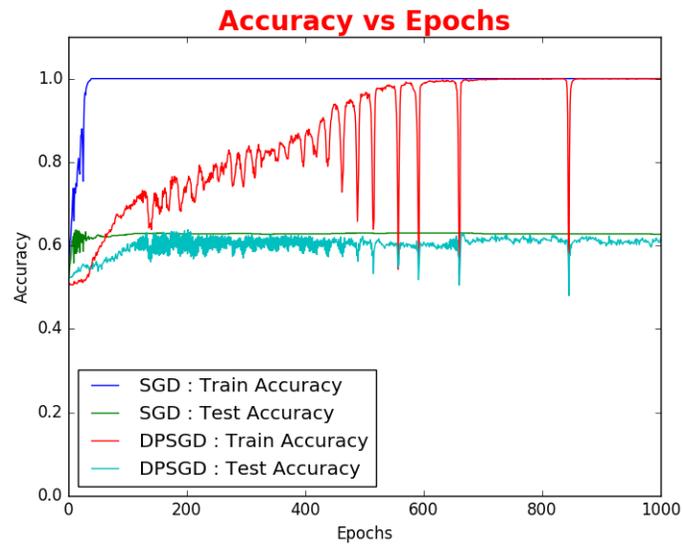

Figure 5.4: Classification accuracy as a function of epochs ($\sigma = 9.5$)

## 5.4 Overfitting and Convergence Analysis

Figures 5.1, 5.2, 5.3 and 5.4 illustrate the classification accuracy for SGD and DPSGD as a function of the number of training epochs. The accuracy is plotted on the Y-Axis and Epochs are plotted on the X-Axis. Each figure contains plots for accuracy on test and training sets for both algorithms.

In figure 5.1, we see that non-private training reaches 100% classification accuracy on training set within a few epochs but the test accuracy remains constant fairly around 60%. 100% accuracy to the training data means that the hypothesis produced by SGD converges completely to the distribution of noise and the underlying class distribution combined. This is counterproductive as we would ideally like to fit only to the underlying data distribution. The huge difference in errors at the points of convergence to the training and test data can be viewed as an indicator of overfitting.

In differentially private training, it can be observed that it takes slightly more epochs in comparison to vanilla SGD for the learning mechanism to produce a hypothesis with comparable performance on the fresh test set i.e it takes slightly longer for DPSGD to produce a hypothesis that converges to the underlying distribution. In Figure 5.3, we observe that the hypothesis generated by Stochastic Gradient Descent converge to the train and test sets around the 40-epoch mark. The hypothesis learnt using DPSGD converge to the training and test sets at 730 and 70 epoch marks respectively. The hypothesis generated by DPSGD takes 1725% more epochs to converge to the training data by while taking only 25% more epochs to produce comparable performance on the fresh data set. We can conclude that it takes very long (a few hundred epochs) for the training algorithm to completely fit to the training data or encode the training data within itself, i.e. overfit to the training data due to data reuse.

We observe that DPSGD significantly increases the horizontal gap between the points of convergence to the training data and the test data. We have seen that DPSGD delays convergence to the noisy training distribution by a few hundred epochs. There are signs that DPSGD actually produces better performance on the test set than SGD, when an optimal privacy bound is used. We can observe this in Figure 5.2. This is evidence to assert that DPSGD is a viable option for controlling over-fitting due to data reuse as it slows down the process of convergence to the training data significantly more than the process of convergence to the test data.

## 5.5   Effect of $\sigma$ on generalization

Figures 5.1, 5.2, 5.3 and 5.4 represent the performance vs epochs curves for $\sigma$ values 1.0, 4.0, 8.5 and 9.5 respectively, with all other parameters being kept constant. From these figures, we observe that increasing $\sigma$ values (or lowering $\epsilon$ values) has two effects: i) The convergence of the classifier to the noisy training data requires more epochs. ii) The convergence of the classifier to the underlying distribution (fresh data) requires lesser number of epochs.

Increasing $\sigma$ corresponds to lower $\epsilon$ and vice-versa. High $\sigma$ means low $\epsilon$ (for constant $\delta$) and thus high noise. Low $\sigma$ means high $\epsilon$ (for constant $\delta$) and low noise. The learning mechanism using higher amounts of noise takes longer to converge to the training data. High noise also means low $\epsilon$ (for constant $\delta$) and thus, tighter privacy guarantees. We can observe that tighter the privacy guarantee, the longer the learning algorithm takes to generate a hypothesis that converges to the underlying data distribution. (See Figures 5.1, 5.2, 5.3 and 5.4) Also, the tighter the privacy guarantee, the sooner the learning algorithm takes to generate a hypothesis that converges with the noisy training data.

## 5.6 Privacy Analysis

From Section 2.3.3, each training step in DPSGD is $(\epsilon, \delta)$-differentially private with respect to the lot and $(O(q\epsilon\sqrt{T}), \delta)$-differentially private on the whole. As the number of epochs increase, so does the privacy budget. For a very high number of training epochs, private and non-private training become very similar. This is explained by the notion that a very high privacy budget is equivalent of having no privacy. For a very low number of training epochs, the accuracy results might not be good enough. In Figure 5.2, there are signs that DPSGD actually produces better performance on the test set than SGD, when an optimal privacy bound is used. We can thus see the need for balancing the accuracy-privacy trade-off.

## 5.7 Summary

This chapter evaluates the over-fitting performance of the training algorithm in relation to the number of training epochs. It demonstrates that DPSGD converges significantly slower to the training data in relation to the test data (Section 5.4) and produces slightly better performance on test data than SGD (Figure 5.2).

# CHAPTER 6

# CONCLUSION

In this work, we have demonstrated that using a differential privacy based optimizing mechanism for Deep Neural Networks helps to minimize the maximum differential error between performance on the train and test sets by 60% while exhibiting comparable performance on the test set. It increases the likelihood of getting low differential errors at the expense of high differential errors. We also report an increase of 25% points in the training error while achieving similar performance on the test data.(Sections 4.5, 4.6)

Furthermore, we have demonstrated that DPSGD delays convergence to the noisy training data, requiring 1725% more epochs as SGD while converges fairly quickly to the underlying distribution (requiring only 25% more epochs as DPSGD). We have also observed a marginal improvement in performance on the test set using DPSGD. (Sections 5.4, 5.7)

The creation of a gap between convergence to the noisy training data and the underlying data distribution, reduction in differential error and improved performance on fresh data make this technique a viable approach to controlling overfitting due to data reuse.

# CHAPTER 7
# FUTURE WORK

There have been some attempts towards using differential privacy based network training in a distributed setting[33]. It would be interesting to see the performace of the DPSGD algorithm while training in a distributed manner. There seems to be some promise in averaging the model weights from training runs on similar or different datasets. This algorithm can further be tested by evaluating its performance on a) data of different kinds like images, videos, text, spatial / temporal data b) more advanced deep learning architectures like Convolutional Neural Networks (CNNs), Recurrent / Recursive Neural networks (RNNs), etc. Using a compination of DPSGD and Dropout might be an interesting direction to explore. Exploring the existence of a privacy based hard stop to training might be interesting. Evaluating this algorithms on benchmark datasets like MNIST, CIFAR, etc. could give us an idea of its performance in reference to current state-of-the-art. Last we also look to study the generalizing behaviour of DPSGD with respect to other notions of generalization.[7]

# REFERENCES

[1] Martín Abadi, Ashish Agarwal, Paul Barham, Eugene Brevdo, Zhifeng Chen, Craig Citro, Greg S. Corrado, Andy Davis, Jeffrey Dean, Matthieu Devin, Sanjay Ghemawat, Ian Goodfellow, Andrew Harp, Geoffrey Irving, Michael Isard, Yangqing Jia, Rafal Jozefowicz, Lukasz Kaiser, Manjunath Kudlur, Josh Levenberg, Dan Mané, Rajat Monga, Sherry Moore, Derek Murray, Chris Olah, Mike Schuster, Jonathon Shlens, Benoit Steiner, Ilya Sutskever, Kunal Talwar, Paul Tucker, Vincent Vanhoucke, Vijay Vasudevan, Fernanda Viégas, Oriol Vinyals, Pete Warden, Martin Wattenberg, Martin Wicke, Yuan Yu, and Xiaoqiang Zheng. TensorFlow: Large-scale machine learning on heterogeneous systems, 2015. Software available from tensorflow.org.

[2] Martín Abadi, Andy Chu, Ian Goodfellow, H Brendan McMahan, Ilya Mironov, Kunal Talwar, and Li Zhang. Deep learning with differential privacy. In *Proceedings of the 2016 ACM SIGSAC Conference on Computer and Communications Security*, pages 308–318. ACM, 2016.

[3] Raef Bassily, Adam D. Smith, and Abhradeep Thakurta. Private empirical risk minimization, revisited. *CoRR*, abs/1405.7085, 2014.

[4] Amos Beimel, Hai Brenner, Shiva Prasad Kasiviswanathan, and Kobbi Nissim. Bounds on the sample complexity for private learning and private data release. *Machine learning*, 94(3):401–437, 2014.

[5] Léon Bottou. Large-scale machine learning with stochastic gradient descent. In *Proceedings of COMPSTAT'2010*, pages 177–186. Springer, 2010.

[6] Rich Caruana, Steve Lawrence, and C Lee Giles. Overfitting in neural nets: Backpropagation, conjugate gradient, and early stopping. In *Advances in neural information processing systems*, pages 402–408, 2001.

[7] Rachel Cummings, Katrina Ligett, Kobbi Nissim, Aaron Roth, and Zhiwei Steven Wu. Adaptive learning with robust generalization guarantees. In *Conference on Learning Theory*, pages 772–814, 2016.

[8] Aymeric Damien et al. Tflearn. https://github.com/tflearn/tflearn, 2016.

[9] Cynthia Dwork. Differential privacy: A survey of results. In *International Conference on Theory and Applications of Models of Computation*, pages 1–19. Springer, 2008.

[10] Cynthia Dwork. A firm foundation for private data analysis. *Communications of the ACM*, 54(1):86–95, 2011.

[11] Cynthia Dwork, Vitaly Feldman, Moritz Hardt, Toni Pitassi, Omer Reingold, and Aaron Roth. Generalization in adaptive data analysis and holdout reuse. In *Advances in Neural Information Processing Systems*, pages 2350–2358, 2015.

[12] Cynthia Dwork, Vitaly Feldman, Moritz Hardt, Toniann Pitassi, Omer Reingold, and Aaron Roth. The reusable holdout: Preserving validity in adaptive data analysis. *Science*, 349(6248):636–638, 2015.

[13] Cynthia Dwork, Krishnaram Kenthapadi, Frank McSherry, Ilya Mironov, and Moni Naor. Our data, ourselves: Privacy via distributed noise generation. In *Eurocrypt*, volume 4004, pages 486–503. Springer, 2006.

[14] Cynthia Dwork, Frank McSherry, Kobbi Nissim, and Adam Smith. Calibrating noise to sensitivity in private data analysis. In *TCC*, volume 3876, pages 265–284. Springer, 2006.

[15] Cynthia Dwork, Aaron Roth, et al. The algorithmic foundations of differential privacy. *Foundations and Trends® in Theoretical Computer Science*, 9(3–4):211–407, 2014.

[16] Cynthia Dwork, Guy N Rothblum, and Salil Vadhan. Boosting and differential privacy. In *Foundations of Computer Science (FOCS), 2010 51st Annual IEEE Symposium on*, pages 51–60. IEEE, 2010.

[17] Caglar Gulcehre, Marcin Moczulski, Misha Denil, and Yoshua Bengio. Noisy activation functions. In Maria Florina Balcan and Kilian Q. Weinberger, editors, *Proceedings of The 33rd International Conference on Machine Learning*, volume 48 of *Proceedings of Machine Learning Research*, pages 3059–3068, New York, New York, USA, 20–22 Jun 2016. PMLR.

[18] Richard H. R. Hahnloser, Rahul Sarpeshkar, Misha A. Mahowald, Rodney J. Douglas, and H. Sebastian Seung. Digital selection and analogue amplification coexist in a cortex-inspired silicon circuit. *Nature*, 405(6789):947–951, Jun 2000.

[19] Kevin Jarrett, Koray Kavukcuoglu, Yann LeCun, et al. What is the best multi-stage architecture for object recognition? In *Computer Vision, 2009 IEEE 12th International Conference on*, pages 2146–2153. IEEE, 2009.

[20] Lukasz Kaiser, Aidan N Gomez, Noam Shazeer, Ashish Vaswani, Niki Parmar, Llion Jones, and Jakob Uszkoreit. One model to learn them all. *arXiv preprint arXiv:1706.05137*, 2017.

[21] Shiva Prasad Kasiviswanathan, Homin K Lee, Kobbi Nissim, Sofya Raskhodnikova, and Adam Smith. What can we learn privately? *SIAM Journal on Computing*, 40(3):793–826, 2011.

[22] Steve Lawrence and C Lee Giles. Overfitting and neural networks: conjugate gradient and backpropagation. In *Neural Networks, 2000. IJCNN 2000, Proceedings of the IEEE-INNS-ENNS International Joint Conference on*, volume 1, pages 114–119. IEEE, 2000.

[23] Yann LeCun. Une procédure d'apprentissage pour réseau a seuil asymetrique (a learning scheme for asymmetric threshold networks). In *Proceedings of Cognitiva 85, Paris, France*. 1985.

[24] Yann LeCun, Yoshua Bengio, and Geoffrey Hinton. Deep learning. *Nature*, 521(7553):436–444, 2015.

[25] Vinod Nair and Geoffrey E Hinton. Rectified linear units improve restricted boltzmann machines. In *Proceedings of the 27th international conference on machine learning (ICML-10)*, pages 807–814, 2010.

[26] Arvind Neelakantan, Luke Vilnis, Quoc V Le, Ilya Sutskever, Lukasz Kaiser, Karol Kurach, and James Martens. Adding gradient noise improves learning for very deep networks. *arXiv preprint arXiv:1511.06807*, 2015.

[27] Andrew Y Ng. Feature selection, l 1 vs. l 2 regularization, and rotational invariance. In *Proceedings of the twenty-first international conference on Machine learning*, page 78. ACM, 2004.

[28] Steven J Nowlan and Geoffrey E Hinton. Simplifying neural networks by soft weight-sharing. *Neural computation*, 4(4):473–493, 1992.

[29] Razvan Pascanu, Tomas Mikolov, and Yoshua Bengio. Understanding the exploding gradient problem. *CoRR*, abs/1211.5063, 2012.

[30] Razvan Pascanu, Tomas Mikolov, and Yoshua Bengio. Understanding the exploding gradient problem. *CoRR, abs/1211.5063*, 2012.

[31] Lutz Prechelt. Automatic early stopping using cross validation: quantifying the criteria. *Neural Networks*, 11(4):761–767, 1998.

[32] David E Rumelhart, Geoffrey E Hinton, Ronald J Williams, et al. Learning representations by back-propagating errors. *Cognitive modeling*, 5(3):1, 1988.

[33] Reza Shokri and Vitaly Shmatikov. Privacy-preserving deep learning. In *Proceedings of the 22nd ACM SIGSAC conference on computer and communications security*, pages 1310–1321. ACM, 2015.

[34] Shuang Song, Kamalika Chaudhuri, and Anand D Sarwate. Stochastic gradient descent with differentially private updates. In *Global Conference on Signal and Information Processing (GlobalSIP), 2013 IEEE*, pages 245–248. IEEE, 2013.

[35] Nitish Srivastava, Geoffrey E Hinton, Alex Krizhevsky, Ilya Sutskever, and Ruslan Salakhutdinov. Dropout: a simple way to prevent neural networks from overfitting. *Journal of Machine Learning Research*, 15(1):1929–1958, 2014.

[36] Vladimir Naumovich Vapnik and Vlamimir Vapnik. *Statistical learning theory*, volume 1. Wiley New York, 1998.

[37] Paul John Werbos. Beyond regression: New tools for prediction and analysis in the behavioral sciences. *Doctoral Dissertation, Applied Mathematics, Harvard University, MA*, 1974.